\documentclass[a4paper,11pt, reqno]{amsart}

\usepackage{graphicx}
\usepackage{amsthm}
\usepackage{float}
\usepackage{amsmath,latexsym}
\usepackage{amssymb}
\usepackage{geometry}
\usepackage{amsfonts}
\usepackage{hyperref}
\usepackage[skip=4pt]{subcaption}
\usepackage[ruled]{algorithm2e} %[ruled,vlined, longend]
\usepackage[normalem]{ulem}

\usepackage[dvipsnames]{xcolor}

\def\R{\mathbb{R}}

\newcommand{\dsum}{\displaystyle\sum}

\usepackage{pgfplots}

\newtheorem{rmk}{Remark}[section]

\usepackage{setspace}

\let\origmaketitle\maketitle
\def\maketitle{
  \begingroup
  \def\uppercasenonmath##1{} % this disables uppercasing title
  \let\MakeUppercase\relax % this disables uppercasing authors
  \origmaketitle
  \endgroup
}
\title[]{\large A Mathematical Programming approach to Binary Supervised Classification with Label Noise}

\author[V. Blanco, A. Jap\'on \MakeLowercase{and} J. Puerto]{{\large V\'ictor Blanco$^\dagger$, Alberto Jap\'on$^\ddagger$ and  Justo Puerto$^\ddagger$}\medskip\\
$^\dagger$IEMath-GR, Universidad de Granada\\
$^\ddagger$IMUS, Universidad de Sevilla}

\date{\today}

\begin{document}

\maketitle

\begin{abstract}
In this paper we propose novel methodologies to construct Support Vector Machine -based classifiers that takes into account that label noises occur in the training sample. We propose different alternatives based on solving Mixed Integer Linear and Non Linear models by incorporating decisions on relabeling some of the observations in the training dataset. The first method incorporates relabeling directly in the SVM model while a second family of methods combines clustering with classification at the same time, giving rise to a model that applies simultaneously similarity measures and SVM. Extensive computational experiments are reported based on a  battery of standard datasets taken from UCI Machine Learning repository, showing the effectiveness of the proposed approaches.

\end{abstract}

\section{Introduction}

The primary goal of supervised classification is to find patterns from a training sample of labeled data in order predict the labels of out-of-sample data, in case the possible number of labels is finite.  Among the most relevant applications of classification methods are those related with security, as in spam filtering or intrusion detection. The main difference of these applications with respect to other uses of classification approaches is that malicious adversaries can adaptively manipulate their data to mislead the outcome of an automatic analysis. For instance, spammers often modify their emails by obfuscating words which typically appear in known spam or by adding words which are likely to appear in legitimate emails. Thus, one has to, not only derive a classification rule from a training sample, able to adequately classify out-of-sample data, but also to take into account that some of the labels might be incorrect. Analyzing the vulnerabilities of classifiers and their robustness against attacks, to better understand how their security may be improved, has recently received growing interest from the scientific community. In \cite{bi2005} the authors propose robust alternatives when the features of the training sample observations  are corrupted. On the other hand, \cite{biggio2011} provides an algorithmic approach to handle adversarial modifications of the labels, in case the labels are independently flipped with the same probability, by correcting the kernel matrix.

As it has just been exposed, doubting on the reliability of the labels on the target variable is usual when having suspicions about the possibility of an intentional flip among these labels. However, it is not by far the only case in which one must think about this possibility. Nowadays, it is commonly said that a data scientist spends around an 80$\%$ of his time dealing with collecting and preprocessing data, meanwhile the other 20$\%$ is used to model and extract information from databases. Thus, mistakes converted into wrong label assignments are very likely to happen. For instance, data can be wrongly identified at the very beginning of the data collection phase, or code errors can occur when preprocessing a database, leading to a dataset with label noise.

In this paper, we propose a methodology to construct a classification rule by means of an \textit{ad hoc} adaptation of a Support Vector Machine classifier that incorporates the detection and correction of label noises in the dataset.  Support Vector Machine (SVM) is a widely-used methodology in supervised binary classification, firstly proposed by Cortes and Vapnik \cite{cortesvapnik95}. Given a number of observations with their corresponding labels, the SVM technique consists, in its simplest form, of finding an hyperplane in the feature space so that each class belongs to a different half-space maximizing the separation between classes (in a training sample) and minimizing some measure of the misclassifying errors. This problem can be cast within the class of convex optimization and its dual has very good properties that allow one to extend the methodology to construct also nonlinear classifiers. Most of the SVM literature concentrates on binary classification where several extensions are available. One can use different measures for the separation between classes~\cite{BPR18,IkedaMurata05a,IkedaMurata05b}, select important features~\cite{LMR18}, apply regularization strategies~\cite{LMC18,l1svm}, use twin separators \cite{twin}, etc.

One of the main reasons of the success of SVM tools in classification, may be that one can project the original data out onto a higher dimensional space where the separation of the classes can be more adequately performed, and still with the same computational effort that was required in the original problem. This property is the so-called \textit{kernel trick}, and very likely this is one of the reasons that has motivated the successful use of this tool  in a wide range of applications \cite{writing,credit,insurance,cancer,cleveland}.

According to \cite{nalepa18}, three main groups of approaches for dealing with noisy datasets have been already proposed in the literature: (1) Design of algorithms which filter noisy and/or mislabeled vectors from the input data \cite{gm2009,hc2013}; (2) Construction of robust classifiers against noisy labeling \cite{duan18}; and (3) Use of noise models (typically, it is retrieved in parallel with the obtained classifier, and they are finally coupled for a higher-quality classification) \cite{xu2006,xiao2015}.

Our proposal falls within the third group of the above approaches. We provide a method to simultaneously construct the SVM-based classifier and to re-label observations which allows us to obtain separating hyperplanes that would had been impossible to obtain throughout standard SVM and that can report much better results for many different problems.

\begin{figure}[H]
\begin{center}
\fbox{\includegraphics[scale=0.55]{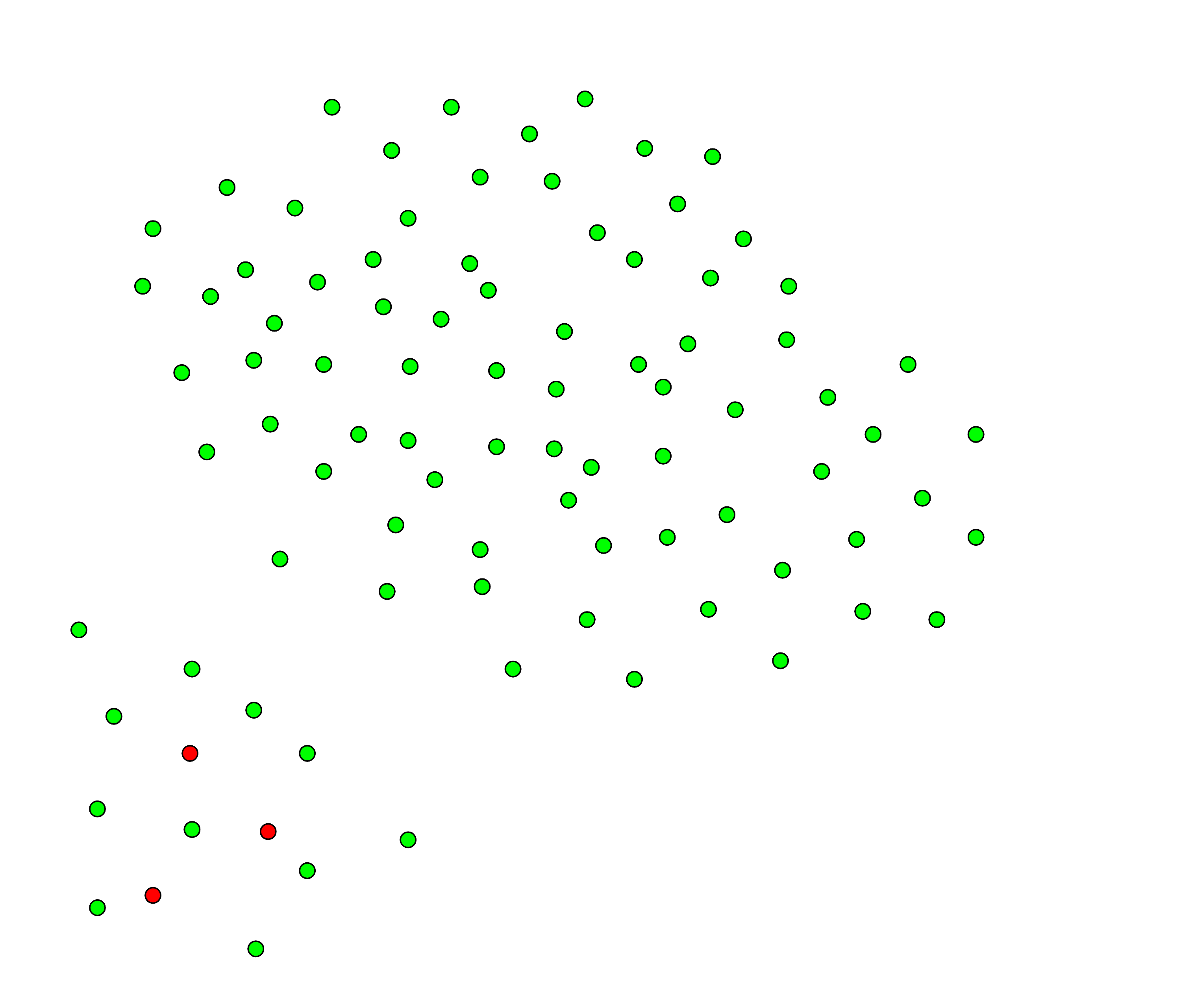}}~\fbox{\includegraphics[scale=0.55]{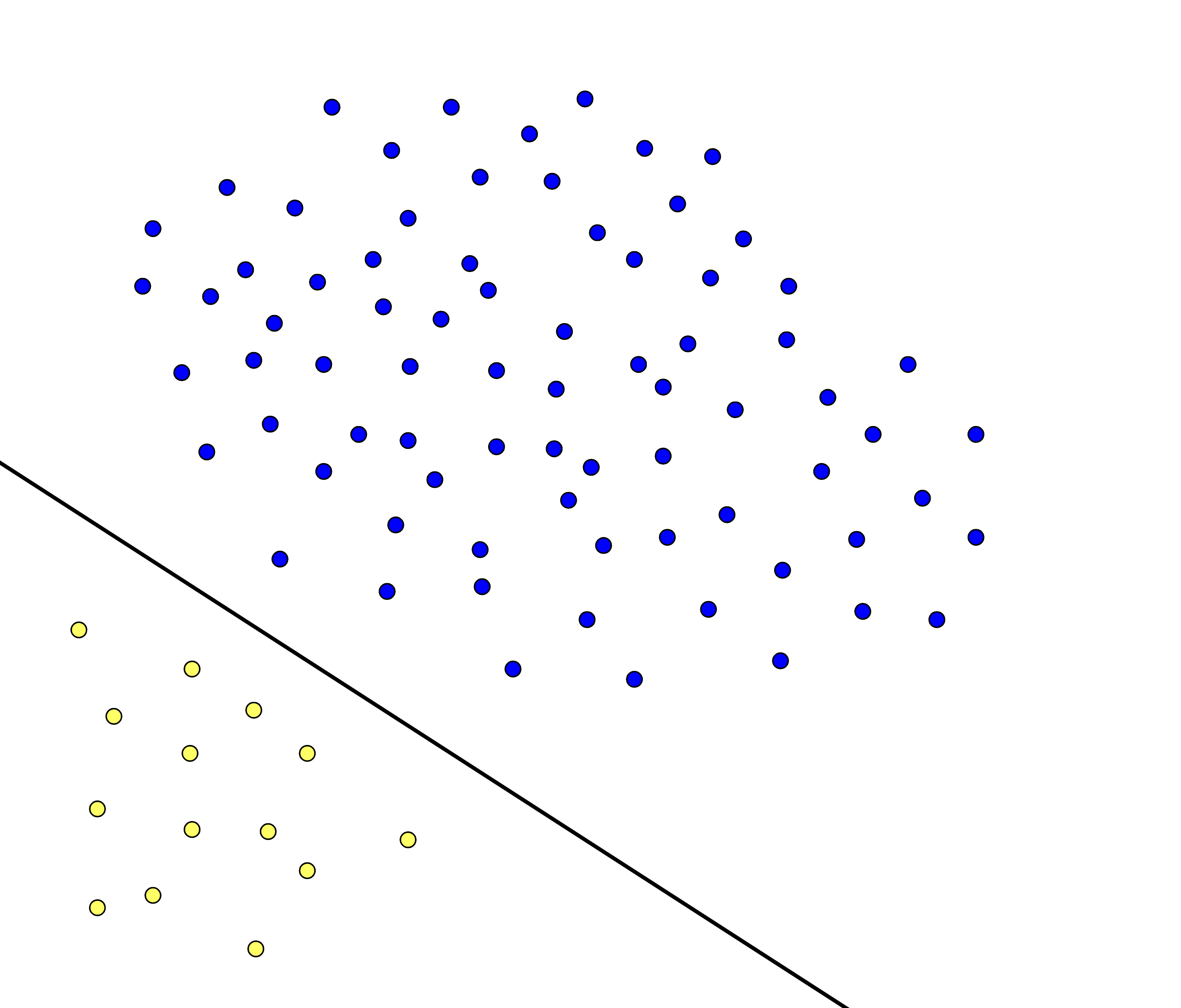}}
\caption{Original data (left) and optimal hyperplane separating re-labeled classes with our method (right).}
\label{fig:1}
\end{center}
\end{figure}

The construction of SVM-based classifiers that simultaneously relabel observations has many advantages when dealing with label noise datasets, but also when working on problems in which false positives and false negatives have different misclassifying costs. Also,  in problems with  unbalanced classes (as for instance in datasets on fraud with credit card transactions in which around a $99.9\%$ of the observations are not fraudulent transactions \cite{FTC17,MBLP17} or in the number of claims in non-life insurances \cite{australiandataset}). In Figure \ref{fig:1} we illustrate this situation. One can observe in the left picture the projection on the plane of a set of observations labeled by fraudulent (red) and non fraudulent (green) transactions. Linear separators seems to be impossible to construct for this instance, but also non linear classifiers will result in overfitting. However, as shown in the right picture, if one allows a few of the labels to be changed, one can obtain better classifiers. Note that in this case, false positives are more costly than false negatives (since asking for a little more of information via text message on the phone normally solves this true negative cases). It is also important to remark that this separating hyperplane could not have been obtained through standard SVM since all the support vectors belong to the same class (green points).

In this paper we propose two different approaches. {We present a model in which re-labeling observations  depends on the errors of the SVM-based method itself searching for a compromise between the gain obtained in misclassification error and margin and the penalty paid for each change of labels}. On the other hand, we will also introduce two models in which re-labeled observations will come from similarity measures on the data.
To asses the validity of these methods we have performed a battery of computational experiments on 6 different real datasets. For these datasets we have repeated the experiments for 5 different scenarios, by randomly flipping a 0$\%$, 20$\%$, 30$\%$, 40$\%$ or 50$\%$ of the labels in the original data. When comparing our method with respect to classical SVM we can see that we obtain better results on noise label datasets.

The rest of the paper is organized as follows. In section 2 we set up and describe the elements of the problem to be considered. Afterward, in section 3 we introduce the different formulations of our models, to end up in section 4 presenting our computational experiments. Finally, we end this article in section 5 with some conclusions and an outline of our future work.

\section{Preliminaries}

In this section we introduce the problem under study and set the notation used through this paper.\\
Given a training sample $\left\lbrace (x_1,y_1),\ldots ,(x_n,y_n)\right\rbrace \subseteq \mathbb{R}^p \times \left\lbrace +1 , -1 \right\rbrace $, the goal a of linear SVM is to obtain a hyperplane separating the data $(x \in \mathbb{R}^p)$ into their two different classes $(y \in \left\lbrace +1 , -1 \right\rbrace)$. Among all possible hyperplanes that can obtain such a separation between the classes, SVM looks for the one with maximum margin (maximum distance from classes to the separating hyperplane) while minimizing the misclassification errors. Let us denote by $\mathcal{H}$ a hyperplane in $\mathbb{R}^p$ in the form $\mathcal{H} = \left\lbrace z \in \mathbb{R}^p \ : \ \omega^t z + \omega_0 = 0 \right\rbrace$ for some $\omega\in \mathbb{R}^p$ and $ \omega_0 \in \mathbb{R}$ (the vector $v^t$ is the result of the transpose operator applied to the vector $v \in \mathbb{R}^p$). This hyperplane will induce a subdivision of the data space $\mathbb{R}^p$ into three regions: the $+1$ (positive) half-space $\mathcal{H}^+ = \left\lbrace z \ : \ \omega^tz+\omega_0 >1 \right\rbrace$, the $-1$ (negative) half-space $\mathcal{H}^- = \left\lbrace z \ : \ \omega^tz+\omega_0 < -1 \right\rbrace$ and the strip $\mathcal{S} =  \left\lbrace z \ : -1 \leq \ \omega^tz+\omega_0 \leq 1 \right\rbrace$.  In the SVM model, positive-class observations ($y = +1$) will be forced to lie on the positive half-space, and the same constraint will be imposed for the negative-class ($y = -1$) observations on the negative half-space. When these constraints are violated for an observation, a penalization error is accounted for in the optimization problem. The separation (margin) between classes is computed as the width of the strip $\mathcal{S}$. As mentioned before, the SVM separating hyperplane will be obtained from an equilibrium of maximizing the separation between classes and minimizing these penalization errors. Denoting by $e_i \in \mathbb{R}^+$ the misclassification error of observation $i$, and by $C$ the constant of penalization of these errors, the SVM can be formulated as the following Non Linear Problem (NLP):

\begin{align}
\min & \ \frac{1}{2}\|\omega\|_2^2 + C\dsum_{i=1}^ne_i & \tag{${\rm SVM}$}\\
\mbox{s.t.} & \;\; y_i(\omega^tx_i+\omega_0) \geq 1-e_i   & \forall i = 1 ,\ldots,n  \nonumber \\
&  \omega \in \mathbb{R}^p, \ \omega_0 \in \mathbb{R},  & \nonumber\\
& e_i \in \mathbb{R}^+, & \forall i = 1,\ldots, n. \nonumber
\end{align}

\begin{figure}
\fbox{\includegraphics[scale=0.11]{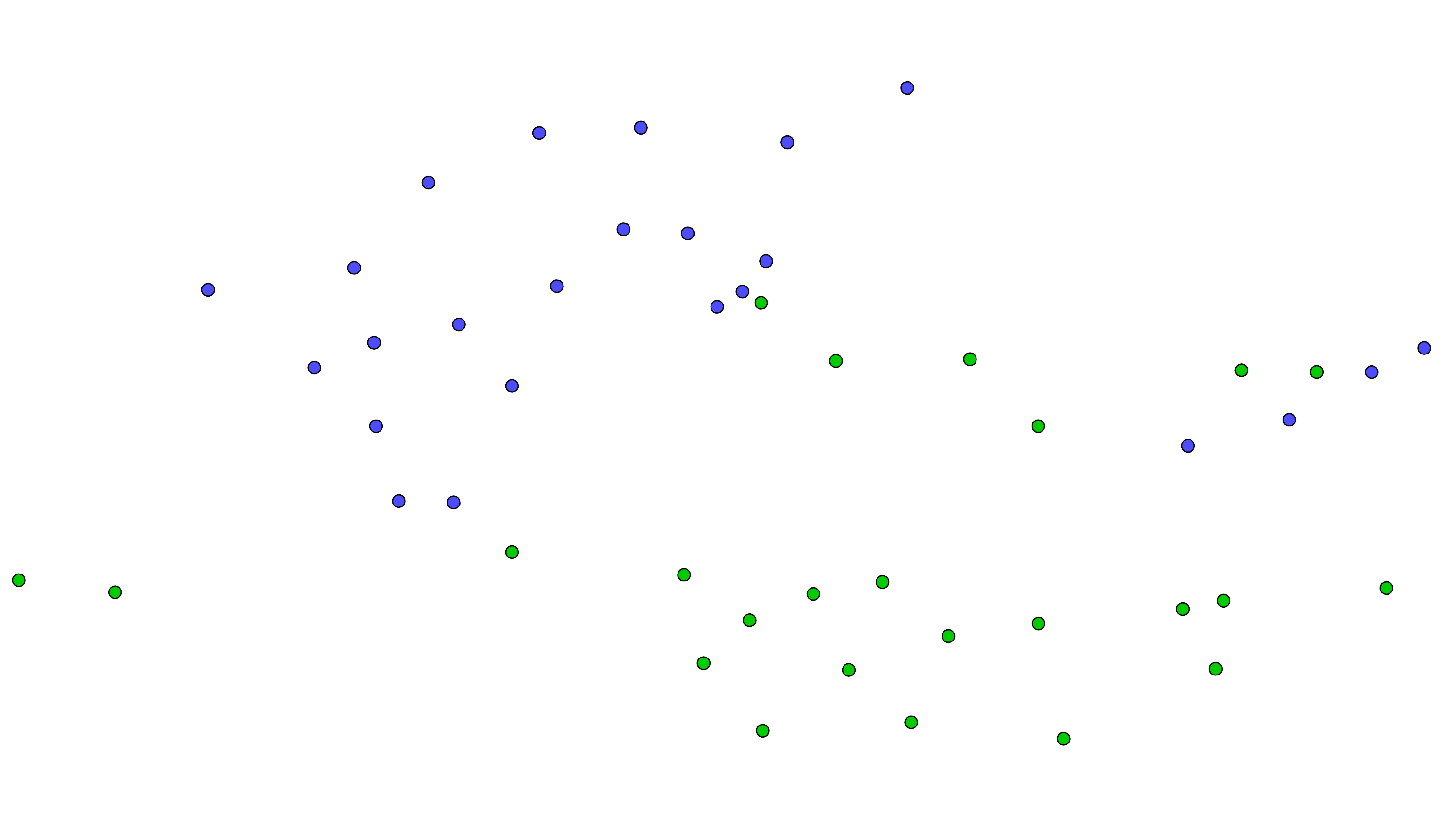}}~
\fbox{\includegraphics[scale=0.11]{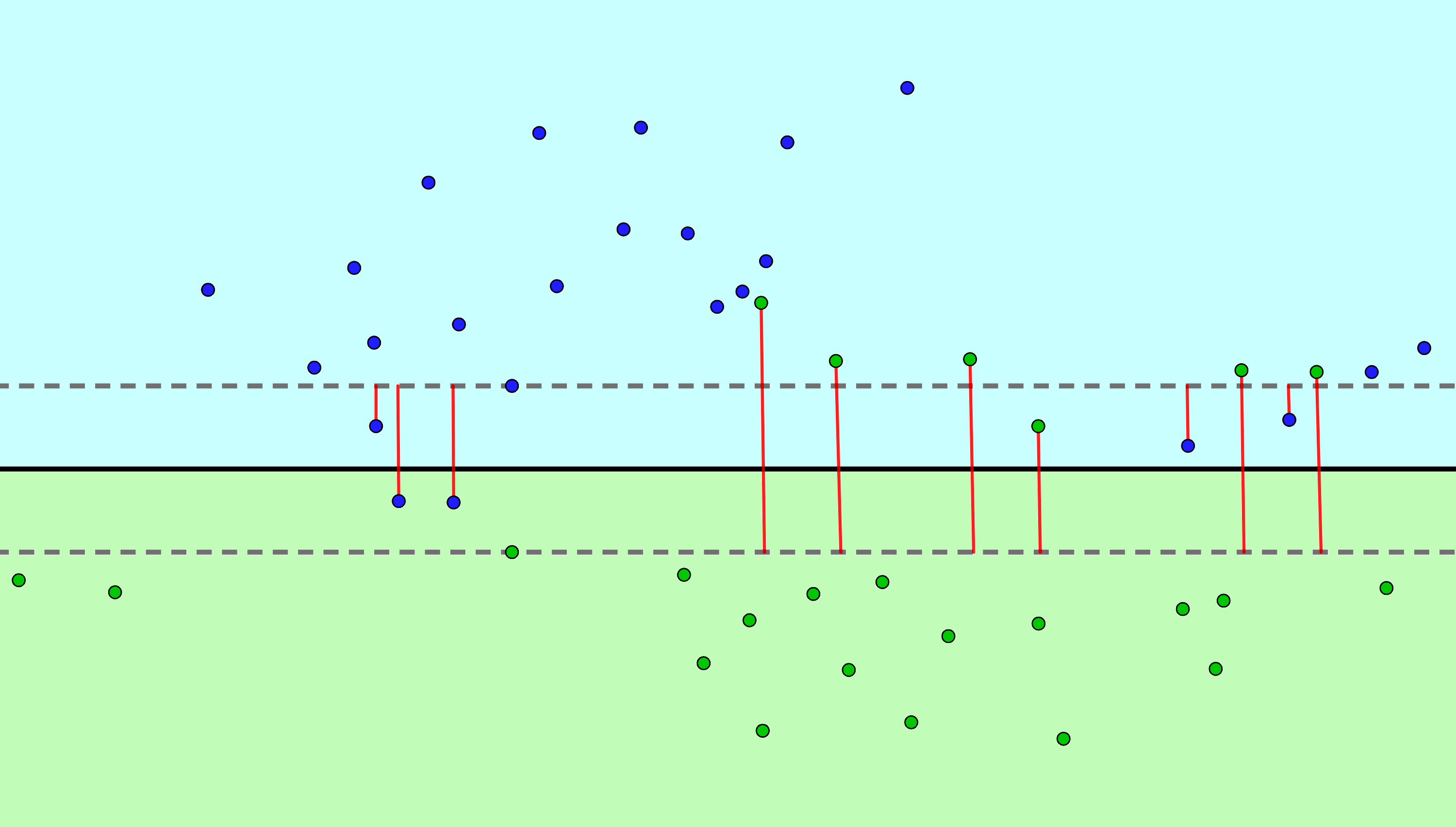}}
\caption{Original set of points (left) and optimal SVM solution on these points (right).}
\label{fig:2}
\end{figure}

In Figure \ref{fig:2} we can see a set of points belonging to two different, blue and green, classes (left picture) and its SVM optimal solution for a given parameter $C$ (right picture). The black line is the separating hyperplane while the other two parallel lines are delimiting the strip, $\mathcal{S}$, between classes. The points that lie on these parallel lines, the boundary of the strip, are the so called support vectors, and they verify that $|\omega^tx_i + \omega_0| =1$. Finally, we represent in red color the magnitude of the errors induced by margin violations.

\begin{figure}[h]
\begin{center}
\includegraphics[scale=0.12]{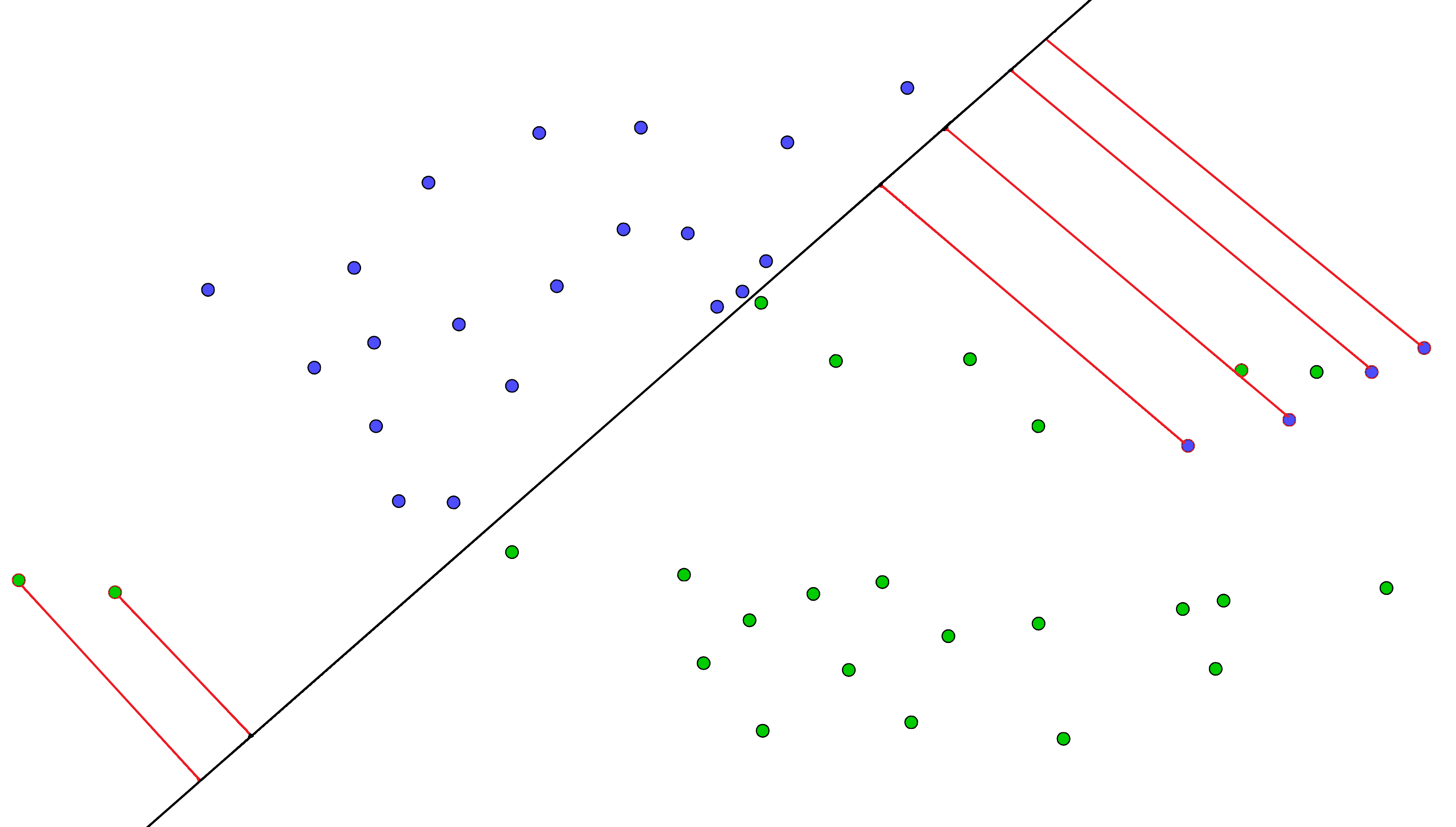}
\caption{Not optimal solution on the SVM problem.}
\label{fig:3}
\end{center}
\end{figure}

If we further analyze the above dataset, we can see that there are four blue observations at the very right of the dataset, and two green observations on the left that have a strong impact when building the classifier. These observations  do not allow one to construct a SVM separator of the dataset as the one we can see in Figure \ref{fig:3}, since that would lead to very big misclassification errors with a very tiny margin.

\begin{figure}[h]
\begin{center}
\includegraphics[scale=0.12]{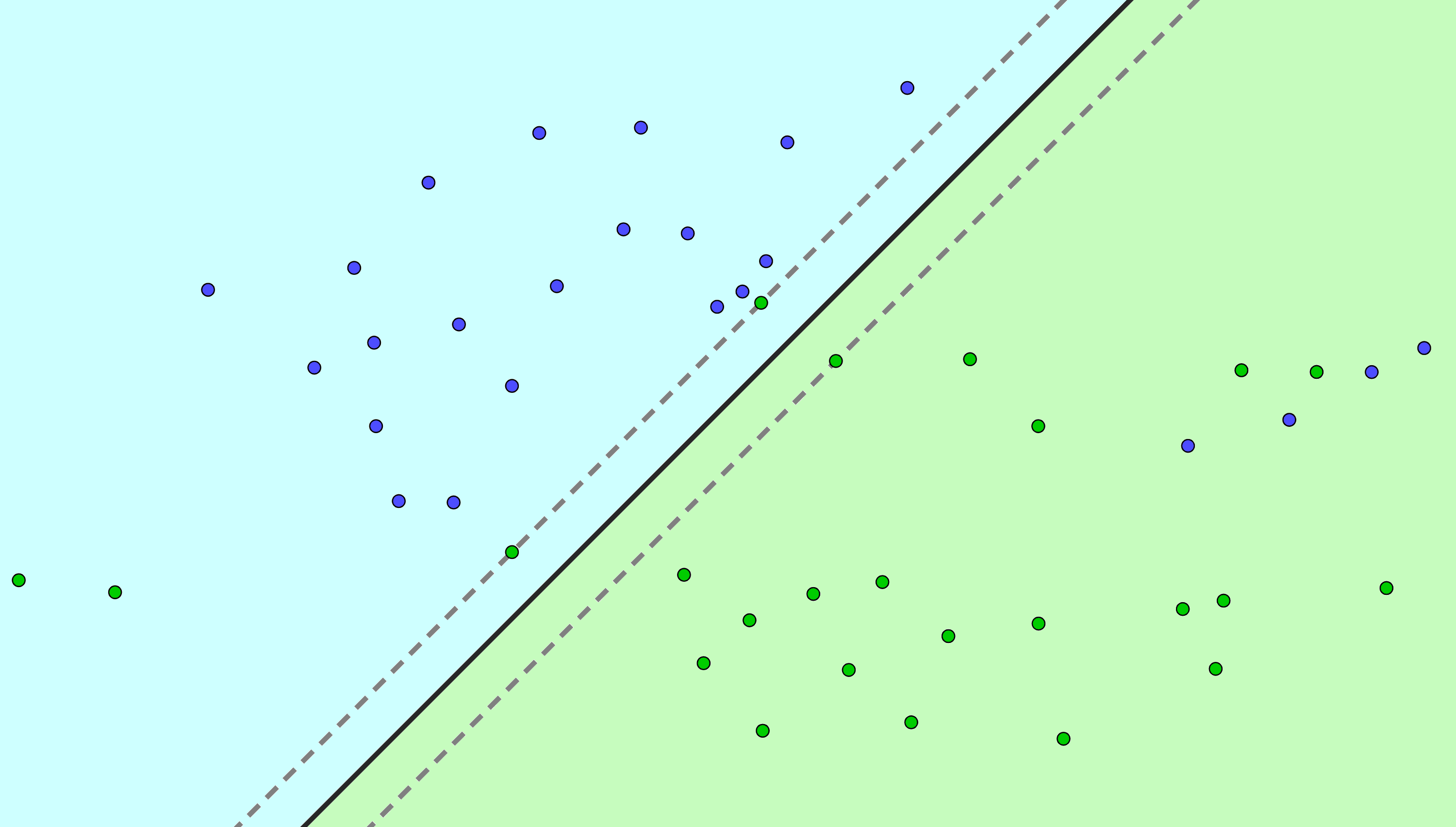}~
\includegraphics[scale=0.12]{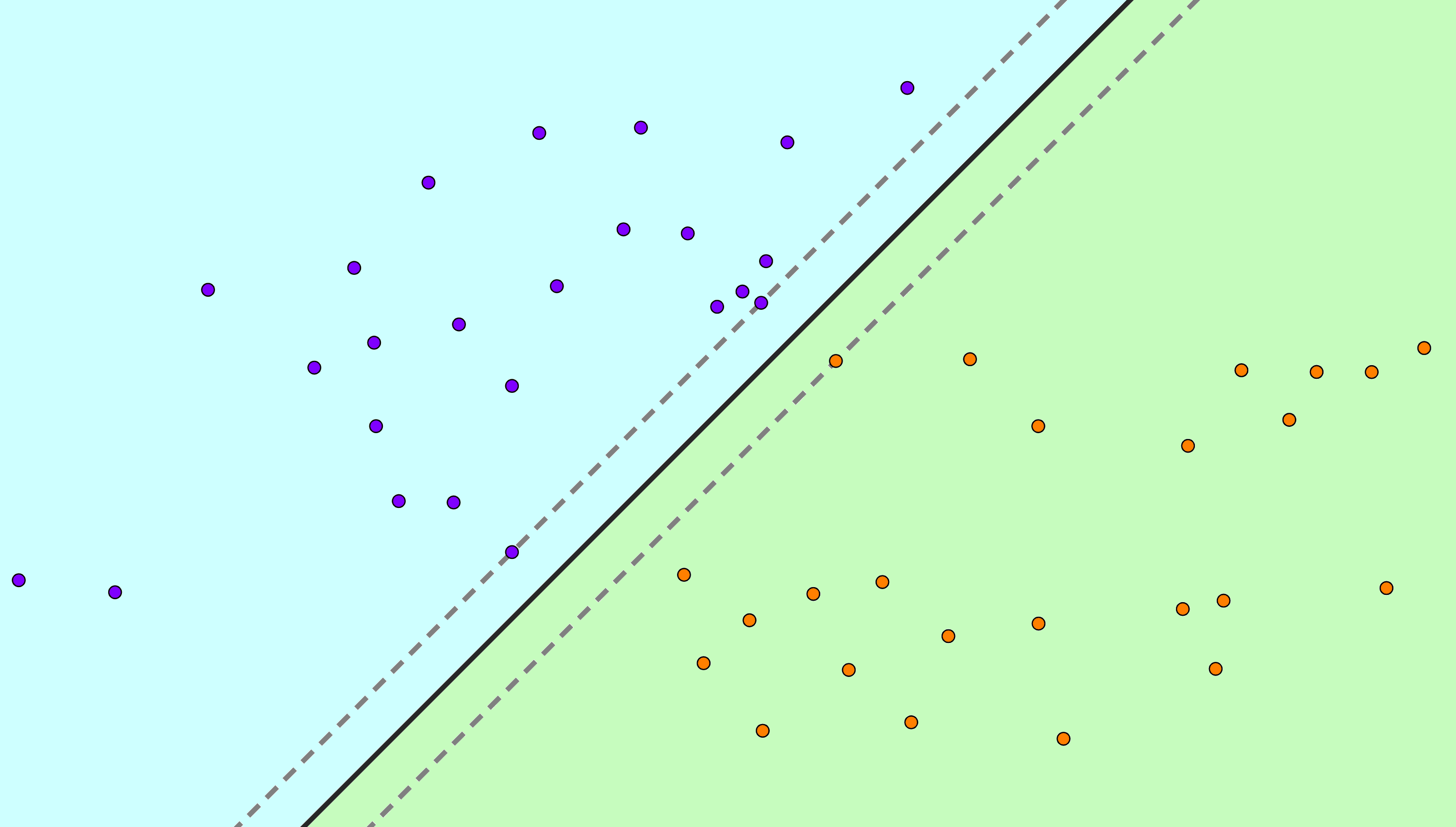}
\caption{Optimal solution after re-labeling.}
\label{fig:4}
\end{center}
\end{figure}

Moreover, there are another two green observations, besides the two on the left, that are closer to the blue cloud of points than to the green one. Hence, if we could consider that these four green points and the four blue ones on the right were wrongly labeled (because of their closeness to the rest of points), we might consider a separating hyperplane with a slope like the one presented on the left of Figure \ref{fig:4} as a better classifier. However, this separating hyperplane would be impossible to obtain with the SVM model since all the support vectors belong to the same class and to avoid huge misclassification errors the model would forbid such a slope.

Motivated by the above kind of configurations, we have studied different models in which a separating hyperplane is obtained not only based on the original labels but also on the possibility of relabeling some of the original observations of the training sample at a given penalty cost. We say that an observation is relabeled if one of the following two assumptions occurs:
\begin{itemize}
\item $y_i = +1$, but our model considers that $y_i=-1$,
\item $y_i = -1$, but our model considers that $y_i=+1$.
\end{itemize}
We will use the notation $\hat{y}_i$ to represent the class that the model is considering for observation $i$. Hence, an observation is said to be relabeled if $ y_i \neq \hat{y}_i$.

Following the example shown in Figures \ref{fig:2} and \ref{fig:3}, we can see on the right of Figure \ref{fig:4} the solution of our model, with a separating hyperplane with the desired slope. Considering the original classes (blue and green), purple points represent the points that the model considers to be blue (despite of their actual label), and orange points represent the points that the model considers to be green. This separating hyperplane is optimal in our problem, the model considers that support points belong to different classes (even thought that is not true regarding to the original values) and no misclassification errors appear in the solution (which is also not true for the original labels).
The underlying idea in these models is that based on the geometry of the problem, relabeling some observations can lead to more robust/accurate classifiers. This classifiers can be very useful when dealing with datasets with outliers, and also in datasets in which some noise is known to be added to the data labels.

\section{Mathematical Programming models}

In this section we present the three mathematical optimization models that we propose to solve the problem consisting on building a hyperplane for binary classification, and, simultaneously, relabeling potential noisy observations. In the first model, relabeling labels on the original observations will be based on the errors with respect to the separating hyperplane. On the other hand, besides considering the errors with respect to the separating hyperplane, the other two models will also take into account information from data based on the geometry of the points through the k-means and the k-medians methods. Nevertheless, despite the fact that some observations are relabeled in our models, in order to make predictions, we will maintain the state for predictions on out of sample data which establishes that observations that lie on the positive half-space of the separating hyperplane will be predicted as positive class observations, meanwhile observations that lie on the negative half-space will be predicted as negative class observations.

\subsection{Model 1: Re-label SVM}$ $\\

The first model that we propose relies on a very basic idea, observations will be relabeled based on the error with respect to the separating hyperplane, i.e., a penalty for each relabeling will be considered and the model will determine whether the cost compensate the global misclassification error. Let $\hat{y}_i$ be the final label for the observation $i$ (after relabeling), for all $i=1,\ldots,n$. Hence, using the notation introduced before, the model can be synthetically summarized in the following way.

\begin{align}
\min \, & \frac{1}{2}\|\omega\|_2^2 + C_1\dsum_{i=1}^ne_i + {\rm Relabeling Cost}(\hat y) & \nonumber\\
\mbox{s.t.} \ \ &  \hat{y}_i(\omega^tx_i+\omega_0) \geq 1-e_i & \forall i = 1 ,\ldots,n \nonumber\\
&  \omega \in \mathbb{R}^p, \ \omega_0 \in \mathbb{R} & \nonumber\\
& e_i \in \mathbb{R}^+ & \forall i = 1,\ldots, n \nonumber,\\
& \hat{y}_i \in \{-1,1\} & \forall i = 1,\ldots, n. \nonumber
\end{align}

The model above is a SVM model in which observations can be relabeled, and thus, instead of considering $y_i$ on the separability constraint, the relabeled observations $\hat{y}_i$ are used. In what follows we describe how to incorporate the relabeling to the constraints and the objective function. Observe that if no cost is assumed for relabeling, the model will relabel most of the observations to obtain a null misclassification error, resulting in senseless classifiers. Thus, we model this cost with a penalty, so that the model will try to maintain the original labels on data and it will only relabel observations when a strong gain on the margin or a strong minimization on the errors is produced.

In order to derive a suitable mathematical programming formulation for the problem, we consider the following set of binary variables to model relabelings:
\begin{center}
$
\xi_i = \left\lbrace
\begin{array}{cc}
1, & \textit{if}\ \ \hat{y}_i = -y_i , \\
0, & \textit{otherwise}.
\end{array}
\right.$ for $i=1, \ldots, n$.
\end{center}

With these variables, $ {\rm Relabeling Cost}(\hat y) = C_2\dsum_{i=1}^n\xi_i$, where $C_2$ is the unitary cost of relabeling. Also, to construct the classifier, we consider the following auxiliary set of continuous variables:
\begin{center}
$
\beta_{ij} = \left\lbrace
\begin{array}{cc}
\omega_j, & \textit{if observation $i$ is relabeled}, \\
0, & \textit{otherwise}.
\end{array} \in \R
\right.$ for $i=1, \ldots, n$, for $j=0, \ldots, p$.
\end{center}
and by $\beta_i= (\beta_{i1}, \ldots, \beta_{ip}) \in \R^p$.

Observe that, with the above notation,
$$
\hat{y}_i (\omega^t x_i + \omega_0) = y_i (\omega^t x_i + \omega_0) - 2 y_i (\beta_i x_i^t +\beta_{i0})
$$

Based on the discussion above, our problem can be formulated as follows:

\begin{align}
\min & \; \frac{1}{2}\|\omega\|_2^2+  C_1 \dsum_{i=1}^n  e_{i}+  C_2\dsum_{i=1}^n\xi_i \label{model:1}\tag{${\rm RE-SVM}$}\\
\mbox{s.t. }\ & y_i(\omega^tx_i + \omega_0) - 2y_i(\beta_i^tx_i + \beta_{i0}) \geq 1-e_i &\forall i = 1 , \ldots , n,\label{c1}\\
& \beta_{ij} = \xi_i\omega_j, &\forall i = 1 , \ldots , n, j=0,\ldots,p,\label{c2}\\
& \omega \in \mathbb{R}^p, \ \omega_0 \in \mathbb{R}, & \label{c3}\\
&  \beta_i\in \mathbb{R}^p, \ \beta_{i0} \in\mathbb{R} & \forall i = 1 , \ldots , n, \label{c3'}  \\
&e_i \in \mathbb{R}^+, \ \xi_i \in \left\lbrace 0, 1 \right\rbrace  & \forall i = 1 , \ldots , n. \label{c4}
\end{align}
In the formulation above, constraints \eqref{c1} and \eqref{c2} allow to model the relabeled observations whereas \eqref{c3} declares that  the coefficients of the hyperplane are continuous variables. Constraint \eqref{c3'} defines a set of variables that will be equal to the coefficients of the hyperplane when an observation is relabeled, and  zero otherwise. With these new coefficients, if an observation is not relabeled, constraints \eqref{c1} coincide with those of the classical SVM, that together with the objective function and \eqref{c4} allow one modeling the misclassification errors as $e_i = \max\{0, 1 - y_i(\omega^tx_i + \omega_0)\}$.

Note that \eqref{model:1} is a Mixed Integer Nonlinear Problem due to its objective function, because even though constraints \eqref{c2} are written in a nonlinear way, they can be linearized as follows:
\begin{align*}
\omega_i - M(1-\xi_i) \leq \beta_{ij}  \leq \omega_i + M(1-\xi_i)\omega_j, & \ \forall i = 1 , \ldots , n, j=0,\ldots,p,\\
-M \xi_i \leq \beta_{ij} \leq M \xi_i, & \ \forall i = 1 , \ldots , n, j=0,\ldots,p.
\end{align*}
for $M \gg 0$ a big enough constant. 

\begin{rmk}
In the same manner that we formulate the problem above using a hinge-loss point of view for the misclassification errors, it can be easily adapted to other loss functions as the ramp loss \cite{RL}. This latter case results in the following mathematical programming model:
\begin{align*}
\min & \frac{1}{2}\|\omega\|_2^2 + C \left( \sum_{i=1}^n e_i+ 2 \sum_{i=1}^n \xi_i \right)& \tag{${\rm RL-SVM}$} \\
\mbox{s.t.} \ \ &  y_i(\omega^tx_i + \omega_0) \geq   1 - e_i - M \xi_i, &  \forall i =1,\ldots,n  \\
&   0 \leq e_i \leq 2, \   & \forall i=1,\ldots,n  \\
&   \xi_i \in \{ 0,1 \},   & \forall i=1,\ldots,n\\
& \omega \in \mathbb{R}^p, \omega_0 \in \mathbb{R}.
\end{align*}
Here, the observations that lie outside the margin in the wrong side of the separating hyperplane are equally penalized in the objective function regardless of the misclassification distance. 
\end{rmk}

\subsection{Cluster-SVM models}

The second family of models that we propose for detecting label noises in the data are based on using similarity measures on the observations. These models will be called  \textit{Cluster-SVM methods} since they perform, simultaneously, two tasks: clustering and classification by SVM. On the one hand, the \textit{cluster} phase of these methods will induce relabeling based on heterogeneity of the information, whereas the SVM phase computes the classifier after relabeling. We present here two different alternatives for clustering data into two groups and its linkage to a classification system: the 2-median and the 2-mean problems.

The goal of these methods is to find two clusters for a given set of observations, considering that an observation will belong to exactly one cluster. These clusters are built by finding two \textit{distinguished points (centroids or medians)} representing each of the two groups determined by the observations closer to them, in a way that the overall sum of distances from points to their respective distinguished points is minimum. We distinguish two models under these settings by using two different distance measures: the $\ell_1$ and the $\ell_2$ norms.

Let us denote by $K_+ \in \R^p$ and $K_- \in \R^p$ the two (unknown) distinguished points, and $d_{i} = \min\{ \|x_i - K_+\|, \|x_i-K_-\|\}$, the distance from the observation $i$ to its closest distinguished points, for $i =1, \ldots, n$  (here $\|\cdot\|$ will represent either the $\ell_1$ or the $\ell_2$-norm). The representation of such a closest distance to the distinguished points will be incorporated to the mathematical programming model using the following set of binary variables:
$$
\theta_i = \left\lbrace
\begin{array}{cc}
1, & \textit{if observation i is assigned to cluster $+$}, \\
0, & \textit{if observation i is assigned to cluster $-$},
\end{array}
\right. \mbox{ for } i=1, \ldots, n.
$$

These clusters represent \textit{similar} observations and will help the SVM methodology, together with the relabeling, to find more accurate classifiers.

Combining the ideas presented on RE-SVM with the clustering based methods, we can derive a new family of models, that assign observations to two groups based on the clusters obtained by minimizing the overall sum of the norm-based distances from the data points to their corresponding reference points. Moreover, it also tries to separate as much as possible these two clusters by means of a hyperplane. Each one of the clusters is assigned to one of the differentiated classes in our classification problem. Finally, this hyperplane will induce a subdivision of the data space in a way that the decision rule of the classification problem for out-of-sample  data is the same that is used in standard SVM. We present below the  MIP formulation for this problem. Let $M\gg 0$ be a big enough positive constant and $\|\cdot\|$ representing either the $\ell_1$ or the $\ell_2$-norm.

\begin{align}
\min & \;\;  \frac{1}{2}\|\omega\| + C_1 \dsum_{i=1}^n  e_{i}+  C_2\dsum_{i=1}^n\xi_i + C_3 \dsum_{i=1}^n d_{i} \label{model_1}\tag{${\rm Cluster-SVM}$}\\\mbox{s.t. }\ & y_i(\omega^tx_i + \omega_0) \geq - M\xi_{i}, &\forall i=1,\ldots,n,\label{c5}\\
& d_{i}\geq \|x_i-K_{+}\| - M(1-\theta_i), &\forall i=1,\ldots,n,\label{c8}\\
& d_{i}\geq \|x_i-K_{-}\| - M\theta_i, &\forall i=1,\ldots,n, \label{c9}\\
& \omega^tx_i + \omega_0 \geq 1 - e_i - M(1-\theta_{i}), &\forall i=1,\ldots,n,\label{c6}\\
& \omega^tx_i + \omega_0 \leq -1 + e_i + M\theta_{i}, &\forall i=1,\ldots,n,\label{c7}\\
& \theta_{i},\ \xi_i \in \{0,1\}, &\forall  i=1,\ldots,n, \label{c10}\\
& e_i, d_{i} \in \mathbb{R}_+, & \forall  i=1,\ldots,n, \label{c11}\\
&  K_+, K_- \in  \mathbb{R}^p, & \label{c12} \\
& \omega \in  \mathbb{R}^p ,\ \omega_0 \in \mathbb{R}.  &\label{c13}
\end{align}

The objective function of \ref{model_1} aggregates the following four elements to be simultaneously optimized:
\begin{itemize}
\item[-] The margin (measured with the $\ell_1$ or $\ell_2$ norm) has to be maximized.
\item[-] The errors of classification with respect to the separating hyperplane have to be minimized.
\item[-] Relabeled observations have to be penalized.
\item[-] Distances from observations to their reference points have to be minimized.
\end{itemize}
The aggregation of these four terms leads to define a hyperplane with a good margin, separating two \textit{homogenous} clusters with respect to distances and classes. Constraint \eqref{c5} enforces the positive (resp. negative) class observations to be located on the positive (resp. negative) half-space of the separating hyperplane. Each relabeled observation is penalized by $C_2$ units, not allowing a large number of relabeling unless it compensates large misclassification errors or unless they lead to a  margin gain. This methodology allows us to keep the same decision rule for out-of-sample data as the one used in standard SVM. Constraints \eqref{c8} and \eqref{c9} permit to determine the closest centroid to each observation, whereas constraints \eqref{c6} and \eqref{c7} enforce the misclassification errors to be computed with respect to the cluster, i.e. the classification is performed with respect to the classes $\hat{y}_i$ that have been created based on the similarity of the observations.

The above model results in two different problems depending on the norm-based distances applied.

\begin{description}
\item[2Median SVM Model] This model results from \eqref{model_1} using the norm  $\ell_1$. It will be referred to as the $2$-Median SVM model. The problem turns out to be a mixed integer linear problem and can be solved using any of the off-the-shelf MIP solvers. 
    %and the efficient strategies which are currently available.
\item[2Mean SVM Model] This is the version of model \eqref{model_1} using the $\ell_2$. Since we are using a nonlinear norm, the $2$-Means SVM results in a Mixed Integer Nonlinear Programming problem, that can be reformulated as a Mixed Integer Second Order Cone Optimization  (MISOCO) problem. As for the MIP there are nowadays available off-the-shelf commercial optimization solvers implementing routines for its efficient solution.
\end{description}

\begin{rmk}[2-$\ell_\tau$ Cluster SVM Model]
One could also consider different $\ell_\tau$-norms ($\tau\geq 1$) for both the margin measure and the clusters similarity measures.
In this case, the problem becomes also a MINLP problem, but based on the results provided in \cite{BPH14}, it can also be efficiently reformulated as a MISOCO problem.
\end{rmk}

\section{Experiments}

In this section we report the results of our computational experience. We have studied six real datasets from UCI Machine Learning Repository (see \cite{biggio2011}), all of them are binary classification problems that come from different topics. The datasets used are: Statlog - Australian Credit Approval (Australian), Breast Cancer (BreastCancer), Statlog - Heart (Heart), Parkinson Dataset with replicated acoustic features (Parkinson), Vertebral Column (Vertebral) and Wholesale Customers (Wholesale). The summarized information about these datasets is detailed in Table \ref{table:1}. For each dataset we report in this table the size ($n$) and the dimension of the problem ($p$).
% Falta poner los nombres no abreviados de los datasets
\begin{table}[h]
\centering
\begin{tabular}{l|cc}
{\bf Dataset} & $n$ & $p$ \\
\hline\hline
\texttt{Australian}  & 690 & 14 \\
\texttt{BreastCancer}  & 683 & 9 \\
\texttt{Heart}  & 270 & 13 \\
\texttt{Parkinson}  & 240 & 40 \\
\texttt{Vertebral}  & 310 & 6 \\
\texttt{Wholesale}& 440 & 7 \\
\hline
\end{tabular}
\caption{Datasets used in our computational experiments.\label{table:1}}
\end{table}

For each of these datasets we have performed five different experiments. The goal in these experiments is to make predictions as accurate as possible on out of sample data. The first experiment consists on making predictions by training the models with the original data. On the other hand, in order to represent attacks in the training data, we have considered four different scenarios in which a random amount of labels, within the set $\left\lbrace 20\%,30\%,40\%,50\% \right\rbrace$, have been flipped for training data, i.e., four scenarios in which we have added some label-noise on training data.\\
We have performed a 5-fold cross validation scheme. Thus, data have been splitted into 5 train-test random partitions. In each of these folds we have trained our models and we have used the other four folds for testing. Moreover, we have repeated this 5-fold cross validation 5 times for each dataset, in order to avoid beneficial starting partitions, and we report the average results obtained. For all the instances we have trained our three models and we have compared them with our benchmark, which is standard SVM. The measure used to evaluate the performance of the models have been the accuracy, in percentage, on out of sample data:
\begin{center}
$\textit{ACC} = \frac{\#\textit{Well Classified Test Observations}}{\#\textit{Test Observations}}\cdot 100$
\end{center}

In each of the instances we have used a grid on the cost parameters and the best result obtained in test among these parameters is the one reported. The grids used in the experiments are the following:

\begin{itemize}
\item[] {\bf SVM:} $C\in \left\lbrace 10^i: i=-5,\ldots , 5 \right\rbrace $.
\item[] {\bf RE-SVM:} $ C_1,C_2 \in \left\lbrace 10^i: i=-5,\ldots , 5 \right\rbrace$.
\item[] {\bf $2$-medians-SVM:} $ C_1,C_2 \in \left\lbrace 10^i: i=-5,\ldots , 5 \right\rbrace $, $C_3 \in \left\lbrace 10^i: i=-3,\ldots , 0 \right\rbrace $.
\item[] {\bf $2$-means-SVM:} $ C_1,C_2 \in \left\lbrace 10^i: i=-5,\ldots , 5 \right\rbrace $, $C_3 \in \left\lbrace 10^i: i=-3,\ldots , 0 \right\rbrace $.
\end{itemize}

\begin{figure}[h]
\begin{center}
\includegraphics[scale=0.45]{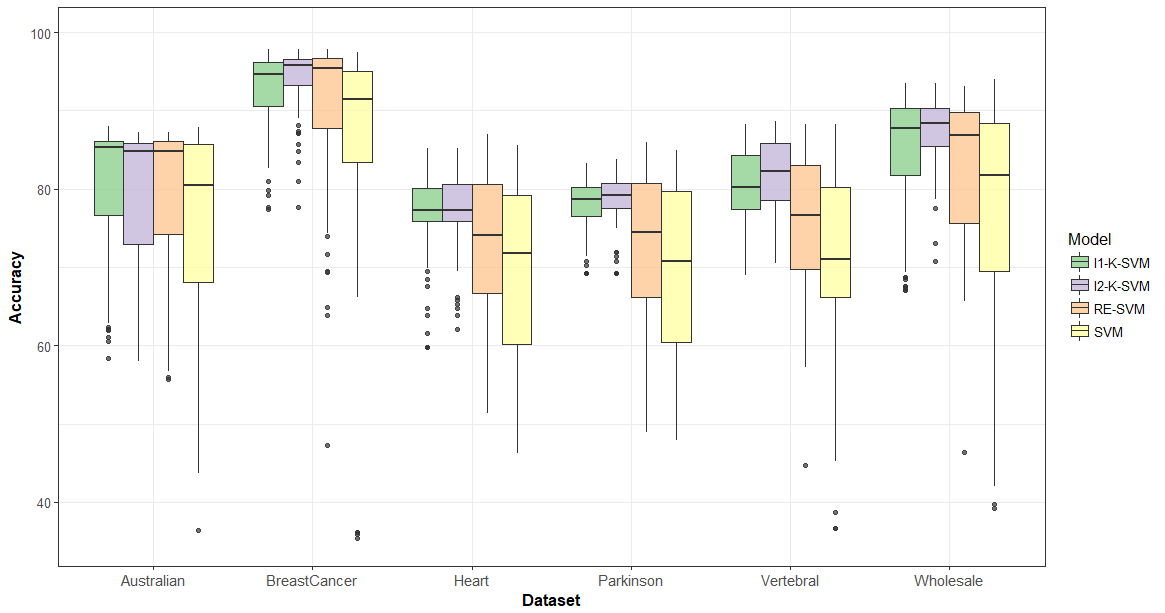}
\caption{Accuracy Boxplots of the obtained accuracies.}\label{fig:5}
\end{center}
\end{figure}

The mathematical programming models were coded in Python 3.6, and solved using Gurobi 7.5.2 on a PC Intel Core i7-7700 processor at 2.81 GHz and 16GB of RAM. We have not solved to optimality all the instances, especially those with the 2-means-SVM in which the problem becomes nonlinear, and hence we have established a time limit of 30 seconds for all the experiments. Moreover, in order to help the solver on the 2-means-SVM, we have upload it with an initial feasible solution that was obtained in the 2-medians-SVM problem.\\
In Table \ref{table:2} we report the average accuracy results obtained in all the experiments for the different models and the different levels of label-noise. In such a table we have used the {\color{SpringGreen} yellow-green color} to indicate the results in which we are a $3\% -5 \%$ better than the benchmark, the {\color{ForestGreen} green color} to indicate whether we are a $5\% -10 \%$ better than the benchmark, and the {\color{Cyan} cyan color} to highlight the results in wich we are at least a $10 \%$ above the benchmark. Regarding to the results, we can conclude that our three models perform better than SVM when an attack in the training data is produced. Besides, the stronger the attack, the bigger the difference between our models' results and SVM's results. We can also point out that 2-medians-SVM and 2-means-SVM perform better than RE-SVM for heavy attacks ($40\% -50\%$ of flipped observations), however, these models require more time to be trained since they have one more hyperparameter to calibrate. To illustrate this, we show in Figure \ref{fig:5} the accuracy boxplots of the 500 instances per dataset (5 partitions $\times$ 5 scenarios $\times$ 5 folds $\times$ 4 models) in which we see how SVM model has lower tails and wider boxes than RE-SVM, and RE-SVM has wider boxes than 2-medians-SVM and 2-means-SVM, which are explained by the behavior of these  models against the attacks.
\begin{table}[H]
\begin{center}
\begin{tabular}{l|l|ccccc}
             &               & 0\% & 20\% & 30\% & 40\% & 50\% \\
             \hline\hline
Australian   & SVM           & 86.11 & 85.43 & 79.23 & 68.13 & 59.47\\
             & RE-SVM        & 86.42 & 85.68 & {\color{SpringGreen}83.37} & {\color{ForestGreen} 76.97} &{\color{ForestGreen}  66.13 } \\
             & 2-medians-SVM & 86.08 & 85.84 & {\color{ForestGreen}84.67} & {\color{Cyan}78.95} & {\color{Cyan}69.54 }\\
             & 2-means-SVM   & 85.97 & 85.74 &{\color{SpringGreen}82.65} & {\color{ForestGreen} 77.14} & {\color{ForestGreen}67.70}\\
             \hline
BreastCancer & SVM           & 96.49 & 93.47 & 89.96 & 85.94 & 68.16\\
             & RE-SVM        & 96.88 & 96.20 & {\color{ForestGreen}94.97} & {\color{SpringGreen} 90.36 } & {\color{ForestGreen} 77.00}\\
             & 2-medians-SVM & 96.63 & 95.31 &{\color{SpringGreen} 94.46} &{\color{ForestGreen} 91.10 }&{\color{Cyan} 87.31 }\\
             & 2-means-SVM   & 96.96 & 95.93 &{\color{ForestGreen} 95.39 }&{\color{ForestGreen} 93.11}& {\color{Cyan} 90.01 } \\
             \hline
Heart        & SVM           & 82.23 & 76.86 & 69.68 & 63.79 & 56.90\\
             & RE-SVM        & 82.84 & 78.38 & {\color{SpringGreen} 73.16} & {\color{ForestGreen}68.86} & {\color{SpringGreen} 61.25}      \\
             & 2-medians-SVM & 82.01 & 78.75 & {\color{ForestGreen} 77.29} & {\color{Cyan} 75.38} & {\color{Cyan} 71.99}\\
             & 2-means-SVM   & 82.06 & 78.81 & {\color{ForestGreen} 77.40} & {\color{Cyan}75.97} & {\color{Cyan}72.90}\\
             		     \hline
Parkinson    & SVM           & 81.66 & 74.74 & 70.17 & 62.28 & 57.82\\		
             & RE-SVM        & 82.43 & 77.64 &{\color{SpringGreen}73.22} &{\color{ForestGreen}67.29}&{\color{ForestGreen}62.97}\\
             & 2-medians-SVM & 80.32 &{\color{SpringGreen}78.62}&{\color{ForestGreen}78.12}& {\color{Cyan}77.51}&{\color{Cyan}76.28}\\
             & 2-means-SVM   & 80.47 &{\color{SpringGreen}79.22}&{\color{ForestGreen}78.78}&{\color{Cyan}78.20}&{\color{Cyan}77.03}\\
             \hline
Vertebral    & SVM           & 84.51 & 75.43 & 71.34 & 66.78 & 57.47\\
             & RE-SVM        & 85.10 & {\color{SpringGreen}79.61} &{\color{SpringGreen} 74.83} & {\color{ForestGreen}72.33 }& {\color{Cyan}67.92}\\
             & 2-medians-SVM & 85.31 & {\color{ForestGreen}82.62} &{\color{ForestGreen} 80.80} & {\color{Cyan}78.30} & {\color{Cyan}76.31}      \\
             & 2-means-SVM   & 86.28 & {\color{ForestGreen}84.32} &{\color{Cyan} 81.77} & {\color{Cyan} 79.91} & {\color{Cyan}76.76}\\
             \hline
Wholesale    & SVM           & 90.08 & 85.30 & 79.74 & 72.23 & 57.73\\
             & RE-SVM        & 90.39 &{\color{SpringGreen} 88.77} &{\color{ForestGreen} 85.97} & {\color{ForestGreen} 80.12} & {\color{Cyan}69.07}\\
             & 2-medians-SVM & 90.58 & {\color{SpringGreen} 89.54} & {\color{ForestGreen}87.79} & {\color{Cyan} 82.78} & {\color{Cyan}73.54}\\
             & 2-means-SVM   & 91.23 & {\color{SpringGreen} 89.56}&{\color{ForestGreen} 87.39} & {\color{Cyan}85.88} & {\color{Cyan}82.92}\\
             \hline \hline
\end{tabular}
\caption{Accuracy results of our computational experiments.}\label{table:2}
\end{center}
\end{table}

\section{Conclusions}

This paper presents a methodology to construct a classification rule  that at the same time incorporates the detection of label noises in the datasets.  Our methodology combines the power of SVM and the features of clustering analysis to simultaneously identify wrong labels to build a separating hyperplane maximizing the margin, minimizing the misclassification errors and penalizing relabeling. The rationale is simple: observations identified as wrongly labeled will be relabeled only if the gain in margin or the decrease in misclassification error compensate the flipping. In spite of its theoretical simplicity we show the exceptional performance of our methodology in a number of databases taken from the UCI repository.

These models are implemented using mathematical programming formulations with some integer variables (MIP). In all cases, they give rise to models that are simple and that enjoy the \textit{quality} of being solvable by nowadays off-the-shelf commercial solver (GUROBI, CPLEX, XPRESS...)

Our findings are not only of theoretical interest. Its practical performance when applied to databases is remarkable. In all tested cases, our methods are superior to the considered benchmark that in our case is standard SVM. Thus, they are directly applicable to datasets in which flipped labels are suspected, resulting in robust classifiers to noisy labels.

Further research on the topic includes, among others, the application of alternative clustering strategies, as those based on ordered median objective functions, the extension of the proposed models to the multiclass  SVM framework or the twin SVM methodology. Also, the the use of kernel tools in our approaches, in order to be able to construct non linear classifiers has to be investigated.

\section*{Acknowledgements}

This research has been partially supported by Spanish Ministry of Education and Science/FEDER grant number  MTM2016-74983-C02-(01-02), and projects FEDER-US-1256951, CEI-3-FQM331 and  \textit{NetmeetData}: Ayudas Fundaci\'on BBVA a equipos de investigaci\'on cient\'ifica 2019.


\begin{thebibliography}{10}
\providecommand{\url}[1]{{#1}}
\providecommand{\urlprefix}{URL }
\expandafter\ifx\csname urlstyle\endcsname\relax
  \providecommand{\doi}[1]{DOI~\discretionary{}{}{}#1}\else
  \providecommand{\doi}{DOI~\discretionary{}{}{}\begingroup
  \urlstyle{rm}\Url}\fi

\bibitem{Agarwal_2018}
Agarwal, N., Balasubramanian, V.N., Jawahar, C.: Improving multiclass
  classification by deep networks using dagsvm and triplet loss.
\newblock Pattern Recognition Letters  (2018).
\newblock \doi{10.1016/j.patrec.2018.06.034}.
\newblock \urlprefix\url{http://dx.doi.org/10.1016/j.patrec.2018.06.034}


\bibitem{writing}
Bahlmann, C., Haasdonk, B., Burkhardt, H.: On-line handwriting recognition with
  support vector machines " a kernel approach.
\newblock In: Proceedings of the Eighth International Workshop on Frontiers in
  Handwriting Recognition (IWFHR'02), IWFHR '02, pp. 49--. IEEE Computer
  Society, Washington, DC, USA (2002).
\newblock \urlprefix\url{http://dl.acm.org/citation.cfm?id=851040.856840}

\bibitem{benders}
Benders, J.F.: Partitioning procedures for solving mixed-variables programming
  problems.
\newblock Numerische mathematik \textbf{4}(1), 238--252 (1962)


\bibitem{bennet-demiriz}
Bennett, K. P.,  Demiriz, A. (1999). Semi-supervised support vector machines. Advances in Neural Information processing systems 11, 368--374.


\bibitem{bi2005} Bi, J.,  Zhang, T. (2005). Support vector classification with input data uncertainty. In Advances in neural information processing systems (pp. 161-168).
\bibitem{biggio2011} Biggio, B., Nelson, B., Laskov, P. (2011, November). Support vector machines under adversarial label noise. In Asian Conference on Machine Learning (pp. 97-112).


\bibitem{BPH14} Blanco, V.,  Ben Ali, S. and Puerto, J. (2014). Revisiting several problems and algorithms in Continuous Location  with $l_p$ norms. Computational Optimization and Applications 58(3): 563-595.

\bibitem{BPS18} Blanco, V., Puerto, J., Salmer\'on, R. (2018). Locating hyperplanes to fitting set of points: A general framework. Computers \& Operations Research, 95, 172-193.

\bibitem{BPR18} Blanco, V., Puerto, J., and Rodr\'iguez-Ch\'ia, A. M. (2020). On $\ell_p $-Support Vector Machines and Multidimensional Kernels. Journal of Machine Learning Research 21 (2020).

\bibitem{BJP18}
Blanco, V., Jap\'on, A., Puerto, J. (2020). Optimal arrangements of hyperplanes for multiclass classification. Advances in Data Analysis and Classification 14, 175–199.

\bibitem{australiandataset} Boucher, J‐P., Denuit, M. and Guillen, M. Number of accidents or number of claims? An approach with zero‐inflated Poisson models for panel data. Journal of Risk and Insurance 76(4), 821--846 (2009).

\bibitem{cortesvapnik95}
Cortes, C., Vapnik, V.: Support-vector networks.
\newblock Machine learning \textbf{20}(3), 273--297 (1995)

\bibitem{knn1}
Cover, T., Hart, P.: Nearest neighbor pattern classification.
\newblock IEEE transactions on information theory \textbf{13}(1), 21--27 (1967)

\bibitem{duan18} Duan, Y., Wu, O. (2018). Learning with auxiliary less-noisy labels. IEEE transactions on neural networks and learning systems, 28(7), 1716-1721.

\bibitem{FTC17} Federal Trade Commission. Consumer sentinel network data book for January-December 2016. March 2017.

\bibitem{ghaddar18} Ghaddar, B., and Naoum-Sawaya, J. (2018). High dimensional data classification and feature selection using support vector machines. European Journal of Operational Research, 265(3), 993-1004.

\bibitem{gm2009} Ghoggali, N., Melgani, F. (2009). Automatic ground-truth validation with genetic algorithms for multispectral image classification. IEEE Transactions on Geoscience and Remote Sensing, 47(7), 2172-2181.

\bibitem{hc2013} Han, X.,  Chang, X. (2013). An intelligent noise reduction method for chaotic signals based on genetic algorithms and lifting wavelet transforms. Information Sciences, 218, 103-118.

\bibitem{credit}
Harris, T.: Quantitative credit risk assessment using support vector machines:
  Broad versus narrow default definitions.
\newblock Expert Systems with Applications \textbf{40}(11), 4404--4413 (2013)


\bibitem{horn} Horn, D., Demircioglu, A., Bischl, B., Glasmachers, T., and Weihs, C. (2016). \emph{A comparative study on large scale kernelized support vector machines}. Advances in Data Analysis and Classification, 1-17.

\bibitem{RL}  Huang, X.L., Shi, L. and Suykens,  J.A.K.
Ramp loss linear programming support vector machine
J. Mach. Learn. Res., 15 (2014), 2185-2211.

\bibitem{IkedaMurata05a} K.~Ikeda and N.~Murata (2005). \emph{Geometrical Properties of Nu Support Vector Machines with Different Norms}. Neural Computation 17(11), 2508-2529.

\bibitem{IkedaMurata05b} K.~Ikeda and N.~Murata (2005). \emph{Effects of norms on learning properties of support vector machines}. ICASSP (5), 241-244


\bibitem{insurance}
Ka{\v{s}}{\'c}elan, V., Ka{\v{s}}{\'c}elan, L., Novovi{\'c}~Buri{\'c}, M.: A
  nonparametric data mining approach for risk prediction in car insurance: a
  case study from the montenegrin market.
\newblock Economic research-Ekonomska istra{\v{z}}ivanja \textbf{29}(1),
  545--558 (2016)


\bibitem{nb}
Lewis, D.D.: Naive (bayes) at forty: The independence assumption in information
  retrieval.
\newblock In: European conference on machine learning, pp. 4--15. Springer
  (1998)

\bibitem{uci}
Lichman, M.: {UCI} machine learning repository (2013).
\newblock \urlprefix\url{UCI Machine Learning Repository}

\bibitem{LMC18} L\'opez, J., Maldonado, S., and Carrasco, M. (2018). Double regularization methods for robust feature selection and SVM classification via DC programming. Information Sciences, 429, 377-389.

\bibitem{LMR18} Labb\'e, M., Mart\'inez-Merino, L. I., and Rodr\'iguez-Ch\'ia, A. M. (2018). Mixed Integer Linear Programming for Feature Selection in Support Vector Machine. Discrete Applied Mathematics, https://doi.org/10.1016/j.dam.2018.10.025.

\bibitem{cancer}
Majid, A., Ali, S., Iqbal, M., Kausar, N.: Prediction of human breast and colon
  cancers from imbalanced data using nearest neighbor and support vector
  machines.
\newblock Computer methods and programs in biomedicine \textbf{113}(3),
  792--808 (2014)
\bibitem{labbe14} Maldonado, S., P\'erez, J., Weber, R., Labb\'e, M. (2014). Feature selection for support vector machines via mixed integer linear programming. Information sciences, 279, 163-175.

\bibitem{MBLP17} S. Maldonado, C. Bravo, J. L\'opez, J. P\'erez (2017) Integrated framework for profit-based feature selection and SVM classification in credit scoring.
Decision Support Systems,  104, 113-121.


\bibitem{mangasarian}  Mangasarian, O.L. \emph{Arbitrary-norm separating plane}. Oper. Res. Lett., 24 (1--
2):15--23 (1999).

\bibitem{martinez2000} Mart\'inez, D., Millerioux, G. (2000). Support vector committee machines. European Symposium on Artificial Neural Networks-ESSANN'2000.

\bibitem{e1071} Meyer, D., Dimitriadou, E., Hornik, K., Weingessel, A. and
   Leisch, F.
   \newblock e1071: Misc Functions of the Department of
  Statistics, Probability Theory Group (Formerly: E1071), TU Wien. R package
  version 1.6-8. \url{https://CRAN.R-project.org/package=e1071} (2017)

\bibitem{nalepa18} Nalepa, J.,  Kawulok, M. (2018). Selecting training sets for support vector machines: a review. Artificial Intelligence Review, 1-44.

\bibitem{python} Pedregosa, F., Varoquaux, G. , Gramfort, A., Michel, V., Thirion, B., Grisel, O., Blondel, M.,  Prettenhofer, P., Weiss, R., Dubourg, V., Vanderplas, J., Passos, A., Cournapeau, D., Brucher, M., Perrot, M. and Duchesnay, E.,
\newblock Scikit-learn: Machine Learning in Python. Journal of Machine Learning Research 12, 2825--2830, 2011.

\bibitem{twin} Peng, X, Chen, D. (2018). PTSVRs: Regression models via projection twin support vector machine
Information Sciences 435, 1--14.

\bibitem{l1svm} Peng, X. Xu, D., Kong, L. and Chen, D. (2016). L1-norm loss based twin support vector machine for data recognition
Information Sciences 340–341, 86-103.

\bibitem{cleveland}
Radhimeenakshi, S.: Classification and prediction of heart disease risk using
  data mining techniques of support vector machine and artificial neural
  network.
\newblock In: Computing for Sustainable Global Development (INDIACom), 2016 3rd
  International Conference on, pp. 3107--3111. IEEE (2016)

\bibitem{xiao2015} Xiao, H., Biggio, B., Nelson, B., Xiao, H., Eckert, C., Roli, F. (2015). Support vector machines under adversarial label contamination. Neurocomputing, 160, 53-62.
\bibitem{xu2006} Xu, L., Crammer, K., Schuurmans, D. (2006, July). Robust support vector machine training via convex outlier ablation. In AAAI (Vol. 6, pp. 536-542).


\end{thebibliography}
\end{document}